\documentclass[10pt,twocolumn,letterpaper]{article}

\usepackage[pagenumbers]{cvpr} 

\usepackage{graphicx}
\usepackage{amsmath}
\usepackage{amssymb}
\usepackage{booktabs}
\usepackage{multirow}

\usepackage[pagebackref,breaklinks,colorlinks]{hyperref}

\usepackage[capitalize]{cleveref}
\crefname{section}{Sec.}{Secs.}
\Crefname{section}{Section}{Sections}
\Crefname{table}{Table}{Tables}
\crefname{table}{Tab.}{Tabs.}

\def\cvprPaperID{xxxx} 
\def\confName{CVPR}
\def\confYear{yyyy}

\begin{document}

\title{Semantically Consistent Person Image Generation}

\author{
  Prasun Roy$^{1}$~~~
  Saumik Bhattacharya$^{2}$~~~
  Subhankar Ghosh$^{1}$~~~
  Umapada Pal$^{3}$~~~
  Michael Blumenstein$^{1}$\\
  $^{1}$University of Technology Sydney~~~~~~
  $^{2}$IIT Kharagpur~~~~~~
  $^{3}$ISI Kolkata\\
  {\tt\small \url{https://prasunroy.github.io}}
}

\twocolumn[{%
\renewcommand\twocolumn[1][]{#1}%
\maketitle
\begin{center}
  \centering
  \captionsetup{type=figure}
  \includegraphics[width=\textwidth]{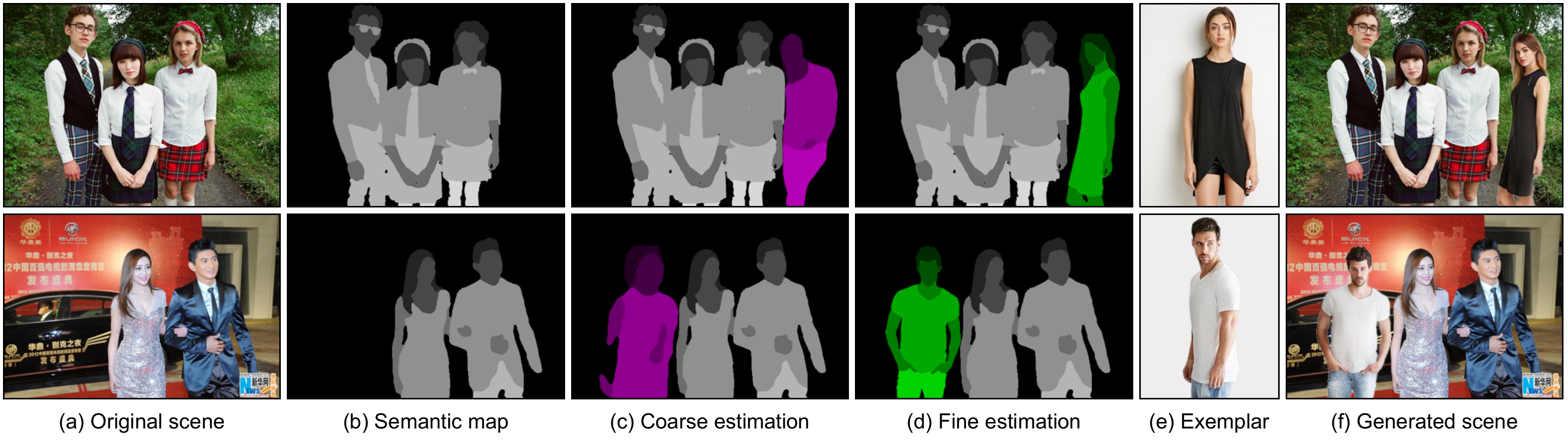}
  \captionof{figure}{Overview of the proposed method. (a) Original scene. (b) Semantic maps of existing persons in the scene. (c) Coarse estimation of the target person's location, scale, and potential pose. (d) Data-driven refinement of the coarse semantic map. (e) An exemplar of the target person. (f) Generated scene with the rendered target person.}
  \label{fig:introduction}
\end{center}%
}]

\begin{abstract}
\vspace{-1.25em}
We propose a data-driven approach for context-aware person image generation. Specifically, we attempt to generate a person image such that the synthesized instance can blend into a complex scene. In our method, the position, scale, and appearance of the generated person are semantically conditioned on the existing persons in the scene. The proposed technique is divided into three sequential steps. At first, we employ a Pix2PixHD model to infer a coarse semantic mask that represents the new person's spatial location, scale, and potential pose. Next, we use a data-centric approach to select the closest representation from a precomputed cluster of fine semantic masks. Finally, we adopt a multi-scale, attention-guided architecture to transfer the appearance attributes from an exemplar image. The proposed strategy enables us to synthesize semantically coherent realistic persons that can blend into an existing scene without altering the global context. We conclude our findings with relevant qualitative and quantitative evaluations.
\end{abstract}

\section{Introduction}\label{sec:introduction}
Person image generation is a challenging yet necessary task for many recent computer vision applications. Though the problem has been majorly addressed by utilizing different generative algorithms, often, the generation quality does not meet the requirements of the practical applications. Moreover, the existing person image generation algorithms rely on two main factors. First, they heavily utilize the appearance and pose attributes of the target to generate the final image \cite{ma2017pose, ma2018disentangled, siarohin2018deformable, esser2018variational, zhu2019progressive, tang2020xinggan, tang2020bipartite}. This approach indirectly demands intricate supervision from the users in the form of keypoints, masks, or text inputs \cite{zhou2019text, roy2022tips}. As these attributes are only associated with the person image being generated, we can assume them as \textit{local attributes} or \textit{local contexts}. Secondly, the generation processes that rely heavily on local contexts often ignore global contextual information like background, camera perspective, or the presence of other people and objects in the scene. These over-simplified generation techniques result in target images that fail to blend into a complex natural scene. In this paper, we have addressed an exciting yet challenging task of person image generation considering the global context of the scene. The proposed method is entirely data-driven and does not require any local context from the user. We circumvent the necessity of user input by estimating the best possible local attributes for the transfer process using the available global attributes. The estimated local attributes are further refined to generate more realistic person images.

The main contributions of the proposed work are as follows.

\begin{itemize}
  \item Unlike most existing methods, the proposed technique considers global attributes to generate person images. Thus, the proposed approach enables us to synthesize human images that can blend into a complex scene with multiple existing persons.
  \item The proposed technique utilizes a data-driven refinement strategy which significantly improves the perceptual quality and visual realism of the generated images.
  \item The data-driven approach provides crude control over the appearance attributes to achieve some extent of generation diversity.
  \item The proposed approach achieves state-of-the-art results in most qualitative and quantitative benchmarks.
\end{itemize}

The rest of the paper is organized as follows. We discuss the relevant literature in Sec. \ref{sec:related_work}. The proposed approach is discussed in Sec. \ref{sec:method}. Sec. \ref{sec:experimental_setup} describes the dataset, experimental protocols, and evaluation metrics. The qualitative and quantitative results are analyzed in Sec. \ref{sec:results}. A detailed ablation study is discussed in Sec. \ref{sec:ablation_study}, followed by an analysis of the limitations of the proposed method in Sec. \ref{sec:limitations}. Finally, we conclude the paper in Sec. \ref{sec:conclusions} with a summary of the major findings, potential use cases, and future scopes.

\section{Related Work}\label{sec:related_work}
Image generation is a complex yet intriguing task in computer vision. Generation of person images under different conditions is particularly important for tasks like pose transfer \cite{neverova2018dense}, virtual try-on \cite{han2018viton}, person re-identification \cite{ye2021deep} etc. With the advancement of Generative Adversarial Networks (GANs), person image generation algorithms have also seen new success. Most work on person image generation focuses on generating a person in a target pose given a source image and target pose attributes. The target pose attributes are given as keypoints \cite{ma2017pose, ma2018disentangled, siarohin2018deformable, esser2018variational, zhu2019progressive, tang2020xinggan, tang2020bipartite}, 3D mask \cite{neverova2018dense}, or text \cite{zhou2019text, roy2022tips}. In \cite{ma2017pose}, the proposed generation framework consists of novel pose synthesis followed by image refinement. An UNet-based model is designed to generate an initial coarse image, which is refined in the second stage by another generative model. In \cite{ma2018disentangled}, the authors propose a two-stage generation algorithm with the help of a multi-branched generation network using the target keypoints. Three mapping functions are learned adversarially to map Gaussian noise to the relevant embedding feature space for targeted manipulation of the generated person image. In \cite{balakrishnan2018synthesizing}, the authors have addressed the generation problem by synthesizing the keypoint-conditioned foreground and the background separately. Zhu et al. \cite{zhu2019progressive} have proposed a keypoint-based pose transfer method by incorporating a progressive attention transfer technique to divide the complex task of the generation into multiple repetitive simpler stages. Researchers have also explored the 3D mask as the conditional attribute in the person image generation pipeline. Li et al. \cite{li2019dense} have estimated dense and intrinsic appearance flow between the poses to guide the pixels during the generation process. In \cite{neverova2018dense}, the authors propose an end-to-end model that incorporates surface-based pose estimation and a generative model to perform the pose transfer task. 

Although several algorithms are proposed for person image generation, they require extensive information about the target pose for the generation process. Moreover, most existing algorithms consider the local attributes in the process, which makes them unsuitable for complex scenes. Recently, in \cite{gafni2020wish}, the authors have considered both local and global attributes for the person insertion problem. While the algorithm exhibits some promising initial results, generating visually appealing scene-aware person images is a largely unexplored problem, with \cite{gafni2020wish} being the only attempt in recent literature to the best of our knowledge.

\section{Method}\label{sec:method}

\begin{figure*}[t]
  \centering
  \includegraphics[width=\linewidth]{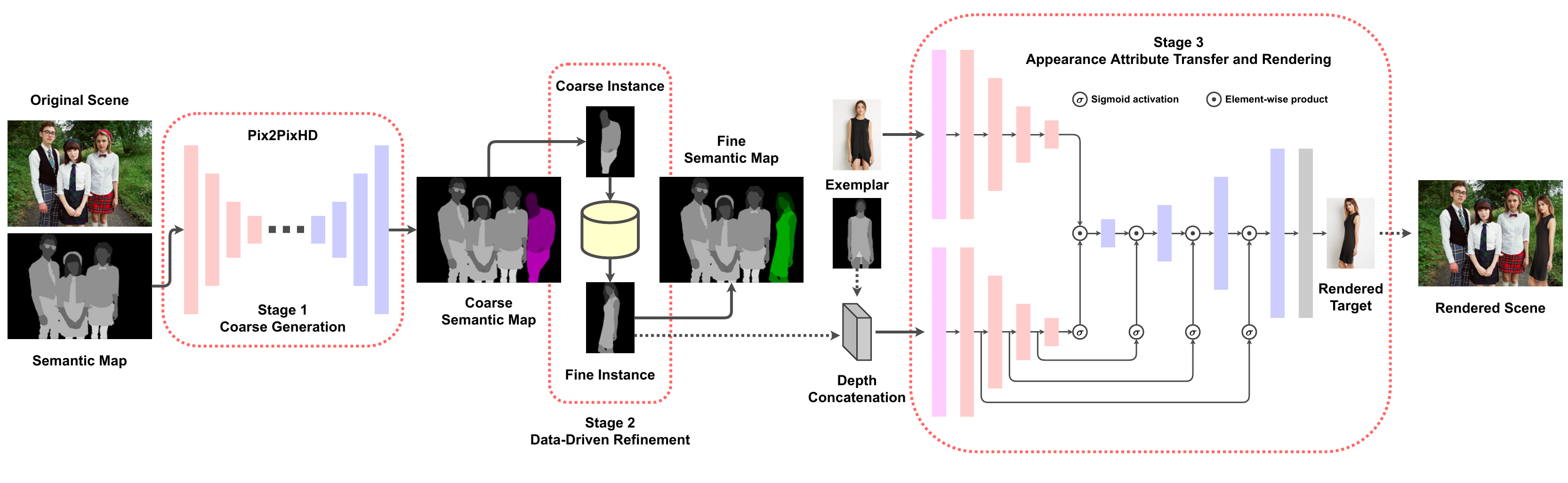}
  \caption{The architecture of the proposed method consists of three main stages. (a) Coarse semantic map estimation from the global scene context in stage 1. (b) Data-driven refinement of the initially estimated coarse semantic map in stage 2. (c) Rendering the refined semantic map by transferring appearance attributes from an exemplar in stage 3.}
  \label{fig:architecture}
\end{figure*}

We propose a three-stage sequential architecture to address the problem. In the first stage, we estimate the potential location and pose of the target person from the global geometric context of the existing persons in the scene. The generated coarse semantic map performs appreciably in providing an estimate of the target location and scale. However, such a crude semantic map performs extremely poorly while attempting to transfer appearance attributes from an exemplar to render the final target. To mitigate this issue, we have taken a data agonistic refinement strategy in the second stage to retrieve a representative semantic map for the target from an existing knowledge base. Finally, we render the target semantic map in the third stage by transferring appearance attributes from an exemplar of the target person. We show an overview of the proposed architecture in Fig. \ref{fig:architecture}.

\subsection{Coarse Generation Network}\label{sec:method_stage1}

\begin{figure}[t]
  \centering
  \includegraphics[width=\linewidth]{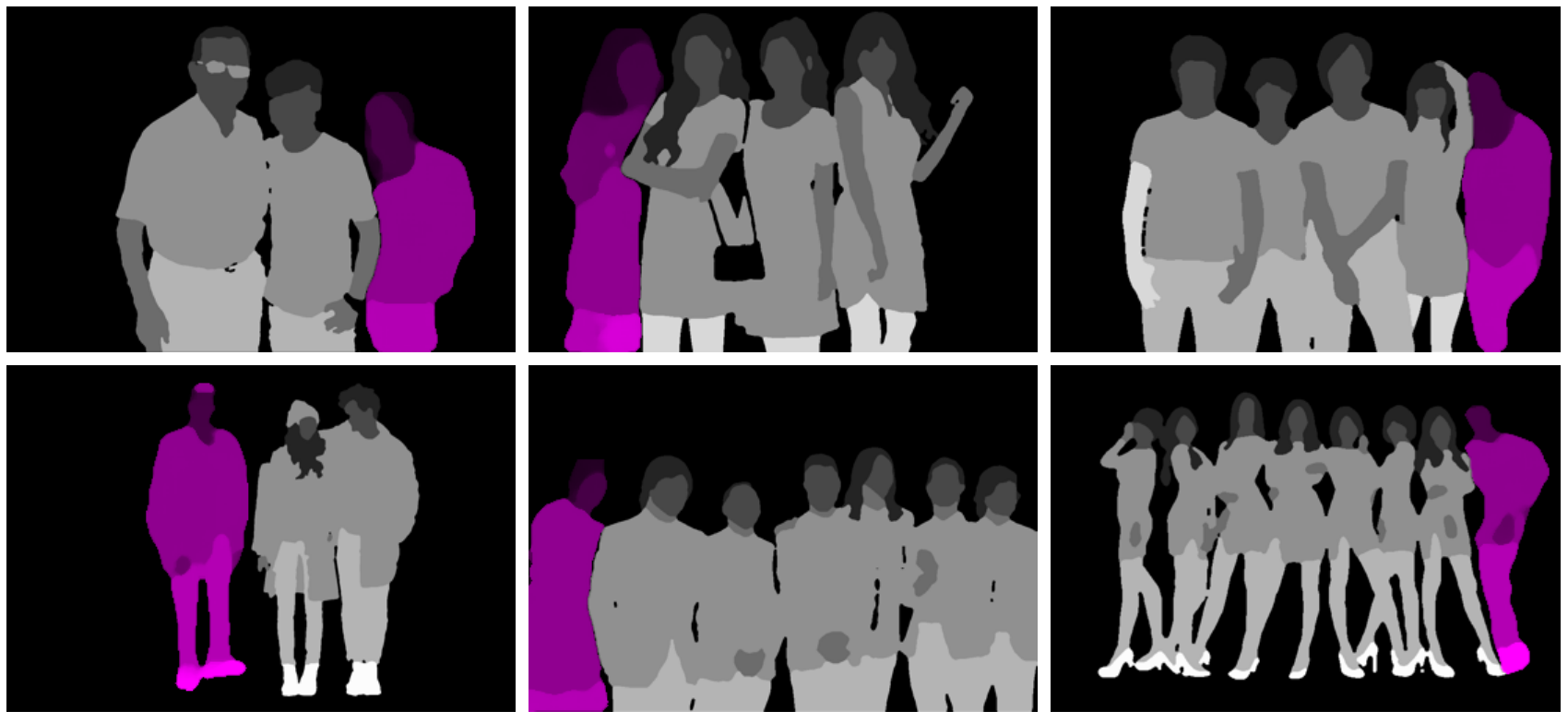}
  \caption{Qualitative results of the coarse generation in stage 1. Semantic maps of existing persons are marked in gray, and the coarse estimation of the target semantic map is marked in purple.}
  \label{fig:result_pix2pixhd}
\end{figure}

We follow a similar approach as \cite{gafni2020wish} to generate a rough estimate of the target person's position, scale and pose. This network performs an image-to-image translation from a semantic map $S$ containing $N$ persons to another semantic map $T$ having the $(N+1)$\textit{-th} person. The network aims to generate a coarse semantic map for a new person such that the new person is contextually relevant to the existing persons in the scene. We show a few examples of the coarse generation network in Fig. \ref{fig:result_pix2pixhd}

Both $S$ and $T$ are single-channel semantic maps containing eight labels corresponding to eight regions of a human body. As mentioned by \cite{gafni2020wish}, this reduced set of label groups simplifies the semantic map generation while retaining sufficient information for high-quality image synthesis in the following stages. The reduced set of semantic label groups contains -- background (0), hair (1), face (2), torso and upper limbs (3), upper body wear (4), lower body wear (5), lower limbs (6), and shoes (7). In \cite{gafni2020wish}, the authors also provide one channel for the face and another optional channel to specify the region boundary for the target. In contrast, we do not consider these additional channels due to our different approaches to refinement and rendering in later stages.

The coarse generation network directly adopts the default encoder-decoder architecture of Pix2PixHD \cite{park2019semantic}. We use a spatial dimension of $368 \times 368$ for the semantic maps. The original semantic maps are resized while maintaining the aspect ratio and then padded with zero to have the desired square dimension. We use nearest-neighbor interpolation when resizing to preserve the number of label groups in the semantic maps. The only modification we apply to the default Pix2PixHD architecture is disabling the VGG feature-matching loss because it is possible to have a wide variation in the target person's location, scale, and pose, which leads to significant uncertainty in the generated semantic map.

\subsection{Data-Driven Refinement Strategy}\label{sec:method_stage2}
The rough semantic map provides a reasonable estimate for the target person, which is contextually coherent with the global semantics of the scene. While the spatial location and scale of the target are immediately usable to localize a new person into the scene, the semantic map itself is not sufficiently viable to produce realistic results. In \cite{gafni2020wish}, authors use a multi-conditional rendering network (MCRN) on the roughly estimated semantic map, followed by a face refinement network (FRN) on the rendered target. While this approach produces some decent results, it is limited in scope due to solely relying on the initially generated rough semantic map from the essence generation network (EGN). We notice two crucial issues in this regard. Firstly, the use of a coarse semantic map highly affects the visual realism of the generated image. Secondly, it is not easy to achieve control over the appearance of the generated target with a fixed semantic representation. For example, if EGN produces a semantic map that appears to be a man while the intended exemplar is a woman. The subtle difference in core appearance attributes between the estimated semantic map and exemplar poses a significant challenge in practically usable generation results. We attempt to improve visual quality and appearance diversity in the generated results by introducing a data-driven refinement strategy with a clustered knowledge base.

We collect a set of finely annotated semantic maps of high-quality human images to construct a small database having a diverse range of natural poses. This database works as a knowledge base for our method. To optimally split the knowledge base into several clusters, we first encode the individual semantic maps using a VGG-19 \cite{simonyan2015very} model pretrained on ImageNet \cite{deng2009imagenet}. The semantic maps are resized to a square grid of size $128 \times 128$, maintaining the aspect ratio and using zero padding. The resampling uses nearest-neighbor interpolation. After passing the resized image through the VGG-19 network, the final feature extraction layer produces an output of dimension $512 \times 4 \times 4$. To avoid too many features during clustering, we apply adaptive average pooling to map the feature space into a dimension of $512 \times 1 \times 1$. The pooled feature space is flattened to a 512-dimensional feature vector. We perform K-means clustering on the encoded feature vectors corresponding to the samples in the knowledge base. From our ablation study in Sec. \ref{sec:ablation_study}, we have found 8 clusters work best for our case. After the algorithm converges, we split the knowledge base by the algorithm-predicted class labels.

During refinement, the coarse semantic map is center-cropped and resized to dimension $128 \times 128$, maintaining the aspect ratio. The resampling uses the same nearest-neighbor interpolation as earlier. The resized coarse semantic map is then similarly encoded and passed to the K-means algorithm for inference. After receiving a cluster assignment, we measure the cosine similarity between the encoded coarse semantic map and every sample previously classified as a cluster member. The refinement returns one or more existing samples by the similarity score-based ranking. The retrieved selection acts as the refined semantic map of the target person.

\subsection{Appearance Attribute Transfer and Rendering}\label{sec:method_stage3}
In \cite{gafni2020wish}, the authors train the rendering network on single instances extracted from multi-person images. In contrast, we impose the rendering task as a pose-transfer problem to transfer the appearance attributes conditioned on the pose transformation. Let us assume a pair of images $I_A$ and $I_B$ of the same person but with different poses $P_A$ and $P_B$, respectively. We aim to train the network such that it renders a realistic approximation $\hat{I}_B$ (generated) of $I_B$ (target) by conditioning the pose transformation $(P_A, P_B)$ on the appearance attributes of $I_A$ (exemplar). We represent each pose with a semantic map consisting of 7 label groups -- background (0), hair (1), face (2), skin (3), upper body wear (4), lower body wear (5), and shoes (6). For effective attribute transfer on different body regions, the semantic map $P$ is converted into a 6-channel binary heatmap (0 for the background and 1 for the body part) $H$ where each channel indicates one specific body region. We use a spatial dimension of $3 \times 256 \times 256$ for $I_A$, $I_B$, and $\hat{I}_B$. Consequently, the same for $H_A$ and $H_B$ is $6 \times 256 \times 256$. We adopt a multi-scale attention-based generative network \cite{roy2022multi,roy2022scene} for rendering. The generator $\mathcal{G}$ takes the exemplar $I_A$ and the depth-wise concatenated heatmaps $(H_A, H_B)$ as inputs to produce an estimate $\hat{I}_B$ for the target $I_B$. The discriminator $\mathcal{D}$ takes the channel-wise concatenated image pairs, either $(I_A, I_B)$ (real) or $(I_A, \hat{I}_B)$ (fake), to estimate a binary class probability map for $70 \times 70$ receptive fields (input patches).

The generator $\mathcal{G}$ has two separate but identical encoding pathways for $I_A$ and $(H_A, H_B)$. At each branch, the input is first mapped to a $64 \times 256 \times 256$ feature space by convolution ($3 \times 3$ kernel, stride=1, padding=1, bias=0), batch normalization, and ReLU activation. The feature space is then passed through 4 consecutive downsampling blocks, where each block reduces the spatial dimension by half while doubling the number of feature maps. Each block consists of convolution ($4 \times 4$ kernel, stride=2, padding=1, bias=0), batch normalization, and ReLU activation, followed by a basic residual block \cite{he2016deep}. The network has a single decoding path that upsamples the combined feature space from both the encoding branches. We have 4 consecutive upsampling blocks in the decoder, where each block doubles the spatial dimension while compressing the number of feature maps by half. Each block consists of transposed convolution ($4 \times 4$ kernel, stride=2, padding=1, bias=0), batch normalization, and ReLU activation, followed by a basic residual block. We apply an attention mechanism at every spatial dimension to preserve both coarse and fine appearance attributes in the generated image. Mathematically, for the first decoder block at the lowest resolution, $k = 1$,
\begin{equation}
  I^D_1 = D_1 (I^E_4 \;\odot\; \sigma(H^E_4))
\end{equation}
and for the subsequent decoder blocks at higher resolutions, $k = \{2, 3, 4\}$,
\begin{equation}
  I^D_k = D_k (I^D_{k-1} \;\odot\; \sigma(H^E_{5-k}))
\end{equation}
where, $I^D_k$ is the output from the $k$\textit{-th} decoder block, $I^E_k$ and $H^E_k$ are the outputs from the $k$\textit{-th} encoder blocks of image branch and pose branch respectively, $\sigma$ denotes the \emph{sigmoid} activation function, and $\odot$ denotes the Hadamard product. Finally, the resulting feature space goes through 4 consecutive basic residual blocks, followed by a convolution ($1 \times 1$ kernel, stride=1, padding=0, bias=0) and \emph{tanh} activation to project the feature maps into the final output image $\hat{I}_B$ of size $256 \times 256$.

The generator loss function $\mathcal{L}_\mathcal{G}$ is a combination of three objectives. It includes a pixel-wise $l_1$ loss $\mathcal{L}^\mathcal{G}_1$, an adversarial discrimination loss $\mathcal{L}^\mathcal{G}_{GAN}$ estimated using the discriminator $\mathcal{D}$, and a perceptual loss $\mathcal{L}^\mathcal{G}_{VGG_\rho}$ estimated using a VGG-19 network pretrained on ImageNet. Mathematically,
\begin{equation}
  \mathcal{L}^\mathcal{G}_1 = \left\| \hat{I}_B - I_B \right\|_1
\end{equation}
where $\|.\|_1$ denotes the $l_1$ norm or the mean absolute error.
\begin{equation}
  \mathcal{L}^\mathcal{G}_{GAN} = \mathcal{L}_{BCE} \left( \mathcal{D} (I_A, \hat{I}_B), 1 \right)
\end{equation}
where $\mathcal{L}_{BCE}$ denotes the binary cross-entropy loss.
\begin{equation}
  \mathcal{L}^\mathcal{G}_{VGG_\rho} = \frac{1}{h_\rho w_\rho c_\rho} \sum_{i=1}^{h_\rho} \sum_{j=1}^{w_\rho} \sum_{k=1}^{c_\rho} \left\| \phi_\rho(\hat{I}_B) - \phi_\rho(I_B) \right\|_1
\end{equation}
where $\phi_\rho$ denotes the output of dimension $c_\rho \times h_\rho \times w_\rho$ from the $\rho$\textit{-th} layer of the VGG-19 network pretrained on ImageNet. We incorporate two perceptual loss terms for $\rho = 4$ and $\rho = 9$ into the cumulative generator objective. Therefore, the final generator objective is given by
\begin{multline}
  \mathcal{L}_\mathcal{G} = \text{arg} \min_{G} \max_{D} \;\; \lambda_1 \mathcal{L}^\mathcal{G}_1 \; + \; \lambda_2 \mathcal{L}^\mathcal{G}_{GAN}\\+ \; \lambda_3 \left( \mathcal{L}^\mathcal{G}_{VGG_4} \; + \; \mathcal{L}^\mathcal{G}_{VGG_9} \right)
\end{multline}
where $\lambda_1$, $\lambda_2$, and $\lambda_3$ are the tunable weights for the corresponding loss components.

The discriminator $\mathcal{D}$ is a generic PatchGAN \cite{isola2017image} that operates on $70 \times 70$ receptive fields of the input. It takes the depth-wise concatenated image pairs, either $(I_A, I_B)$ or $(I_A, \hat{I}_B)$, as a real (1) or fake (0) image transition, respectively.

The discriminator loss $\mathcal{L}_\mathcal{D}$ has only a single component $\mathcal{L}^\mathcal{D}_{GAN}$, calculated as the average BCE loss over real and fake transitions. Mathematically,
\begin{equation}
  \mathcal{L}^\mathcal{D}_{GAN} = \frac{1}{2} \left[ \mathcal{L}_{BCE} (\mathcal{D} (I_A, I_B), 1) + \mathcal{L}_{BCE} (\mathcal{D} (I_A, \hat{I}_B), 0) \right]
\end{equation}
Therefore, the final discriminator objective is given by
\begin{equation}
  \mathcal{L}_\mathcal{D} = \text{arg} \min_{D} \max_{G} \;\; \mathcal{L}^\mathcal{D}_{GAN}
\end{equation}

\section{Experimental Setup}\label{sec:experimental_setup}
\textbf{Datasets:} We use the multi-human parsing dataset LV-MHP-v1 \cite{li2017multiple} to train the coarse generation network in stage 1. The dataset contains 4980 high-quality images, each having at least two persons (average is three), and the respective semantic annotations for every individual in the scene. The annotation includes 19 label groups -- background (0), hat (1), hair (2), sunglasses (3), upper clothes (4), skirt (5), pants (6), dress (7), belt (8), left shoe (9), right shoe (10), face (11), left leg (12), right leg (13), left arm (14), right arm (15), bag (16), scarf (17), and torso skin (18). As discussed in Sec. \ref{sec:method_stage1}, we reduce the original label groups to 8 by merging as -- background + bag (0), hair (1), face (2), both arms + torso skin (3), hat + sunglasses + upper clothes + dress + scarf (4), skirt + pants + belt (5), both legs (6), both shoes (7). While training the coarse generation network, we select one random instance of a scene as the target person and the remaining instances as the input context. We prepare 14854 training pairs from 4945 images and 115 test pairs from the remaining 35 images.

For data-driven refinement in stage 2 and rendering network in stage 3, we use the DeepFashion \cite{liu2016deepfashion} dataset. The dataset contains high-quality single-person instances with wide pose and attire variations. A subset of the samples has color annotations for 16 semantic label groups. We reduce the number of label groups to 7 by merging multiple semantic regions as -- background + bag (0), hair + headwear (1), face + eyeglass (2), neckwear + skin (3), top + dress + outer (4), skirt + belt + pants (5), leggings + footwear (6). We prepare 9866 images and corresponding semantic maps for creating our clustered database. We select 9278 image pairs for training and 786 image pairs for testing the rendering network.

\textbf{Training details:} We train the coarse generation network with batch size 16 and VGG feature-matching loss disabled. All other training parameters are kept to defaults as specified by the authors of Pix2PixHD \cite{park2019semantic}.

The clustering follows Lloyd's K-means algorithm with 8 clusters, a relative tolerance of $1e^{-4}$, 1000 maximum iterations, and 10 random initializations for the centroids.

For the rendering network, we set $\lambda_1 = 5$, $\lambda_2 = 1$, and $\lambda_3 = 5$ in the generator objective. The parameters of both the generator and discriminator networks are initialized before optimization by sampling values from a normal distribution of mean = 0 and standard deviation = 0.02. We use the stochastic Adam optimizer \cite{kingma2015adam} to update the parameters of both networks. We set the learning rate $\eta = 1e^{-3}$, $\beta_1 = 0.5$, $\beta_2 = 0.999$, $\epsilon = 1e^{-8}$, and weight decay = 0 for both optimizers. The network is trained with batch size 4.

\textbf{Evaluation metrics:} Although quantifying visual quality is an open challenge in computer vision, researchers widely use a few quantifiable metrics to assess the perceptual quality of generated images. Following \cite{ma2017pose, siarohin2018deformable, esser2018variational, zhu2019progressive, tang2020xinggan, tang2020bipartite, gafni2020wish}, we calculate Structural Similarity Index (SSIM) \cite{wang2004image}, Inception Score (IS) \cite{salimans2016improved}, Detection Score (DS) \cite{liu2016ssd}, PCKh \cite{andriluka20142d}, and Learned Perceptual Image Patch Similarity (LPIPS) \cite{zhang2018unreasonable} for quantitative benchmarks. SSIM considers image degradation as the perceived change in the structural information. IS estimates the KL divergence \cite{kullback1951information} between the label and marginal distributions for many images using the Inception network \cite{szegedy2015going} as an image classifier. DS measures the visual quality as an object detector's target class recognition confidence. PCKh quantifies the shape consistency based on the fraction of correctly aligned keypoints.

\section{Results}\label{sec:results}

\begin{figure*}[t]
  \centering
  \includegraphics[width=\linewidth]{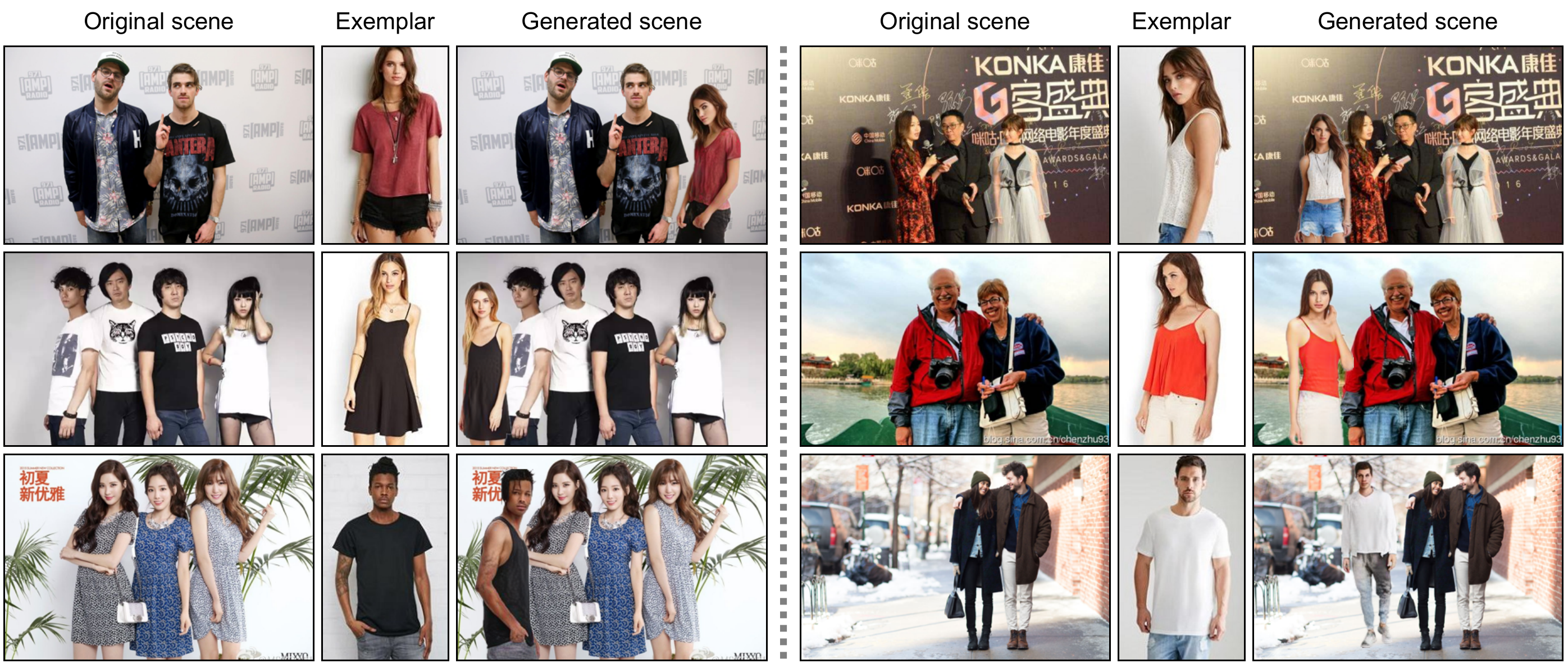}
  \caption{Qualitative results generated by the proposed method. Each set of examples shows -- the original scene (\textbf{left}), an exemplar of the target person (\textbf{middle}), and the final generated scene (\textbf{right}).}
  \label{fig:result_final}
\end{figure*}

\begin{figure}[t]
  \centering
  \includegraphics[width=\linewidth]{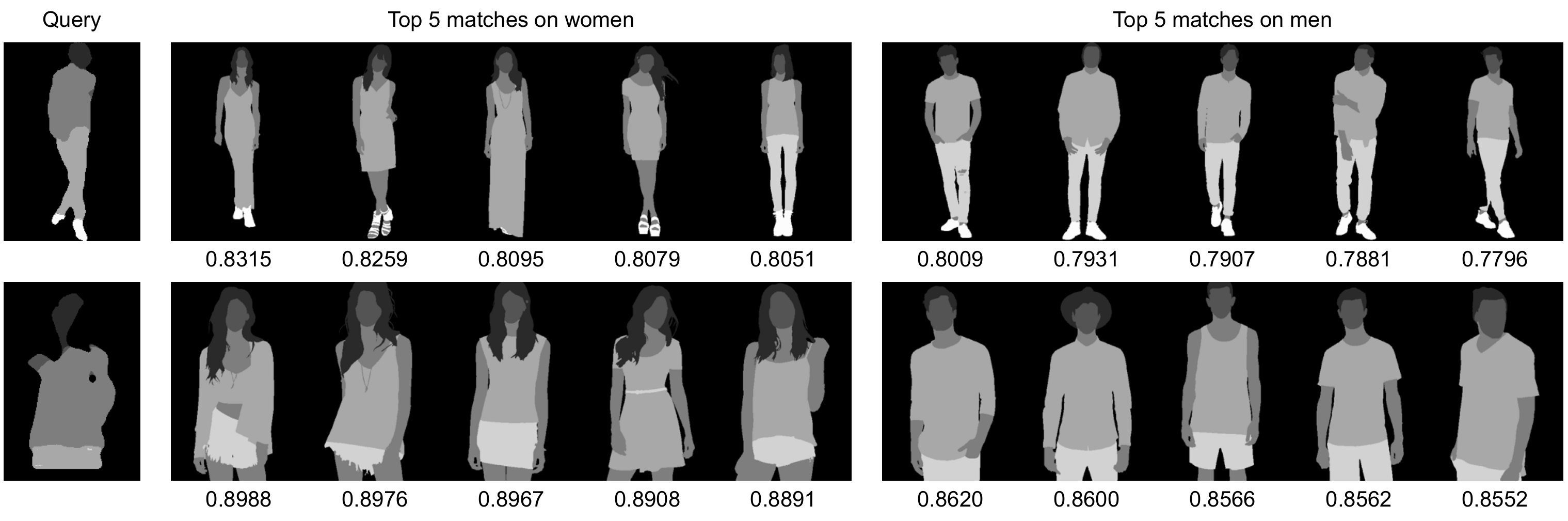}
  \caption{Qualitative results of refinement in stage 2. The first column shows a coarse semantic map as the query, and the following columns show the top-5 refined semantic maps retrieved for both genders. The cosine similarity score for each retrieval is shown below the respective sample. (Best viewed with 400\% zoom)}
  \label{fig:result_clustering}
\end{figure}

\begin{table}[t]
\centering
\caption{Quantitative comparison of the rendering network with existing methods.}
\label{tab:eval_comparison}
\resizebox{\columnwidth}{!}{%
\begin{tabular}{l|cccccc}
\hline
Method &
SSIM $\uparrow$ &
IS $\uparrow$ &
DS $\uparrow$ &
PCKh $\uparrow$ &
\begin{tabular}[c]{@{}c@{}}LPIPS $\downarrow$\\ (VGG)\end{tabular} &
\begin{tabular}[c]{@{}c@{}}LPIPS $\downarrow$\\ (SqzNet)\end{tabular} \\ \hline
$\text{PG}^2$ \cite{ma2017pose}      & 0.773 & 3.163 & 0.951 & 0.89 & 0.523 & 0.416 \\
Deform \cite{siarohin2018deformable} & 0.760 & 3.362 & 0.967 & 0.94 & -     & -     \\
VUNet \cite{esser2018variational}    & 0.763 & \textbf{3.440} & 0.972 & 0.93 & -     & -     \\
PATN \cite{zhu2019progressive}       & 0.773 & 3.209 & \textbf{0.976} & 0.96 & 0.299 & 0.170 \\
XingGAN \cite{tang2020xinggan}       & 0.762 & 3.060 & 0.917 & 0.95 & 0.224 & 0.144 \\
BiGraphGAN \cite{tang2020bipartite}  & 0.779 & 3.012 & 0.954 & 0.97 & 0.187 & 0.114 \\
WYWH (KP) \cite{gafni2020wish}       & 0.788 & 3.189 & -     & -    & 0.271 & 0.156 \\
WYWH (DP) \cite{gafni2020wish}       & 0.793 & 3.346 & -     & -    & 0.264 & 0.149 \\
Ours                                 & \textbf{0.845} & 3.351 & 0.968 & \textbf{0.97} & \textbf{0.124} & \textbf{0.064} \\ \hline
Real Data                            & 1.000 & 3.687 & 0.970 & 1.00 & 0.000 & 0.000 \\ \hline
\end{tabular}%
}
\end{table}

We have performed an extensive range of experiments to explore and analyze the effectiveness of the proposed framework. In Fig. \ref{fig:result_final}, we show a few qualitative results for person image insertion. The final modified scene containing a synthesized person is generated from the original scene and a given exemplar of the target person. It is important to note that no local attribute about the final rendered scene is provided to the generator. To analyze the overall generation quality of the rendering network, we perform a quantitative comparison against eight recently proposed person image generation algorithms \cite{ma2017pose, siarohin2018deformable, esser2018variational, zhu2019progressive, tang2020xinggan, tang2020bipartite, gafni2020wish}. As shown in Table \ref{tab:eval_comparison}, the proposed rendering method outperforms existing algorithms in most evaluation metrics. One of our method's main contributions is refining the initially estimated coarse semantic map to achieve highly detailed person image generation. As we perform a nearest-neighbor search in the semantic feature space of samples in pre-computed clusters, given a coarse semantic map, we can dynamically select a refined candidate for either \emph{women} or \emph{men} as per requirements. This step can be automated if the gender of the exemplar is either known or estimated using a trained classifier. In Fig. \ref{fig:result_clustering}, we show top-5 matches for both \emph{women} and \emph{men} samples given a coarse semantic map as the query to the cluster.

\section{Ablation Study}\label{sec:ablation_study}

\begin{figure*}[t]
  \centering
  \includegraphics[width=\linewidth]{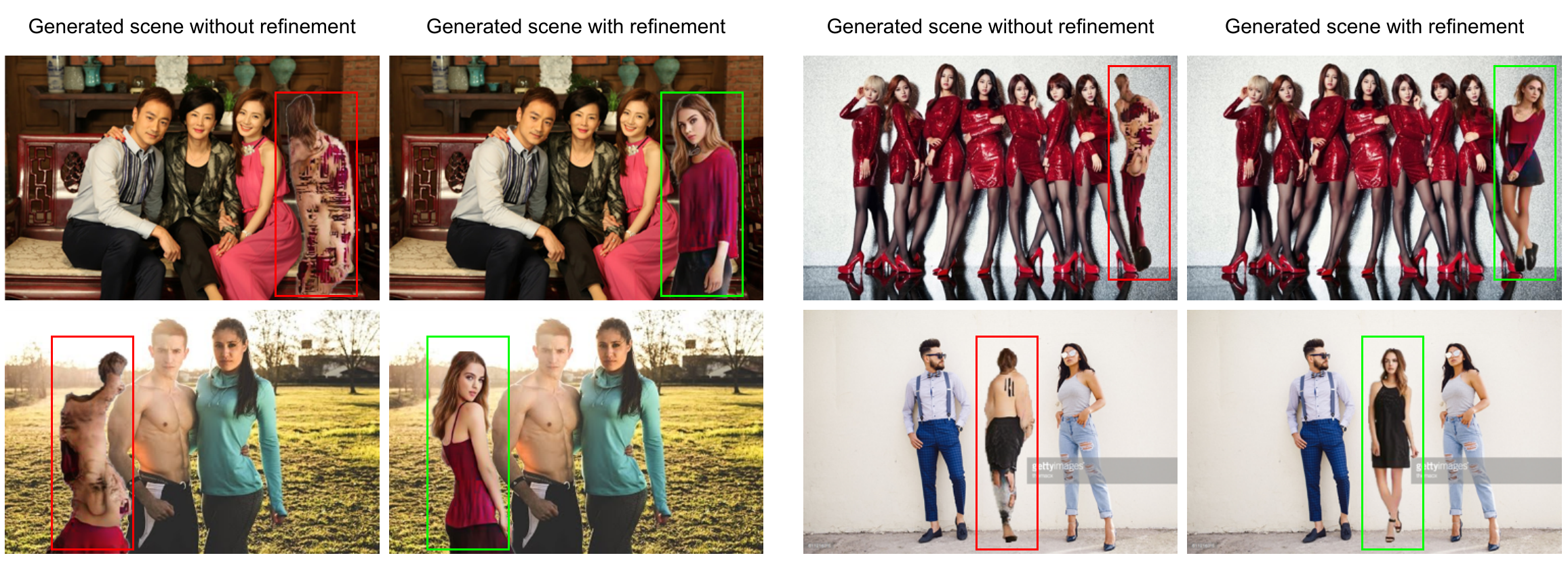}
  \caption{Effects of the refinement strategy on rendering. Each pair of examples show a rendered human in the modified scene \emph{without} (\textbf{left}) or \emph{with} (\textbf{right}) the refinement, marked with red and green bounding boxes, respectively.}
  \label{fig:ablation_refinement}
\end{figure*}

\begin{figure}[t]
  \centering
  \includegraphics[width=\linewidth]{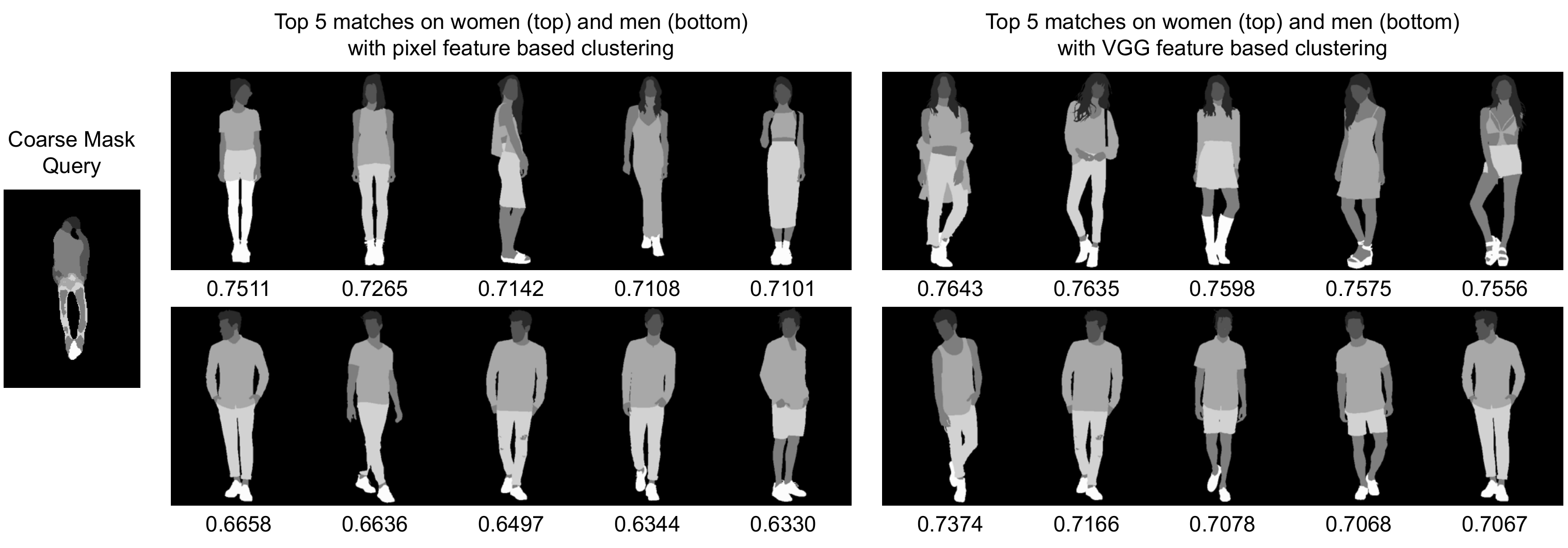}
  \caption{Visual ablation study on the feature representation for clustering. The cosine similarity score for each retrieval is shown below the respective sample. (Best viewed with 400\% zoom)}
  \label{fig:ablation_clustering}
\end{figure}

\begin{figure}[t]
  \centering
  \includegraphics[width=\linewidth]{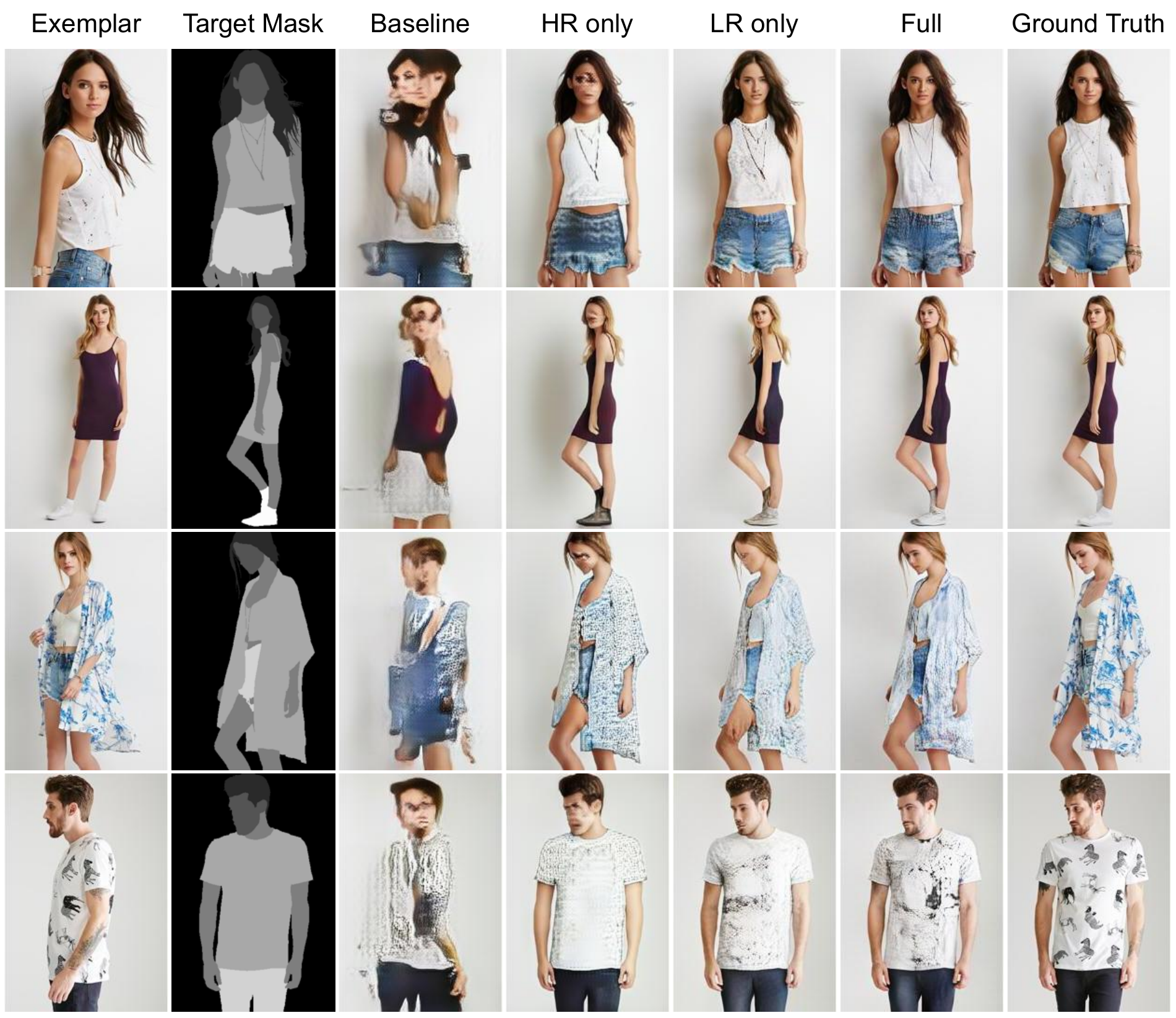}
  \caption{Visual ablation study of the rendering network.}
  \label{fig:ablation_rendering}
\end{figure}

\begin{table}[t]
\centering
\caption{Ablation study of clustering with VGG-encoded features.}
\label{tab:ablation_clustering_vgg}
\resizebox{\columnwidth}{!}{%
\begin{tabular}{l|ccc|ccc}
\hline
\multirow{2}{*}{Number of clusters} & \multicolumn{3}{c|}{Average cosine similarity of top match $\uparrow$} & \multicolumn{3}{c}{Average cosine similarity of top 5 matches $\uparrow$} \\ \cline{2-7} 
       & Men    & Women  & Overall & Men    & Women  & Overall \\ \hline
K = 8  & \textbf{0.8212} & \textbf{0.8319} & \textbf{0.8390}  & 0.7933 & \textbf{0.8171} & \textbf{0.8245}  \\
K = 16 & 0.8184 & 0.8307 & 0.8371  & \textbf{0.7941} & 0.8146 & 0.8227  \\
K = 32 & 0.8073 & 0.8313 & 0.8379  & 0.7824 & 0.8140 & 0.8225  \\
K = 64 & 0.7995 & 0.8290 & 0.8368  & 0.7715 & 0.8109 & 0.8208  \\ \hline
\end{tabular}%
}
\end{table}

\begin{table}[t]
\centering
\caption{Ablation study of clustering with pixel features.}
\label{tab:ablation_clustering_pixel}
\resizebox{\columnwidth}{!}{%
\begin{tabular}{l|ccc|ccc}
\hline
\multirow{2}{*}{Number of clusters} & \multicolumn{3}{c|}{Average cosine similarity of top match $\uparrow$} & \multicolumn{3}{c}{Average cosine similarity of top 5 matches $\uparrow$} \\ \cline{2-7} 
       & Men    & Women  & Overall & Men    & Women  & Overall \\ \hline
K = 8  & 0.7127 & \textbf{0.7562} & \textbf{0.7608}  & 0.6912 & \textbf{0.7366} & \textbf{0.7402}  \\
K = 16 & \textbf{0.7146} & 0.7539 & 0.7598  & \textbf{0.6933} & 0.7357 & 0.7402  \\
K = 32 & 0.7014 & 0.7449 & 0.7492  & 0.6768 & 0.7270 & 0.7302  \\
K = 64 & 0.5852 & 0.6767 & 0.6810  & 0.5580 & 0.6301 & 0.6346  \\ \hline
\end{tabular}%
}
\end{table}

\begin{table}[t]
\centering
\caption{Ablation study of the rendering network.}
\label{tab:ablation_rendering}
\resizebox{\columnwidth}{!}{%
\begin{tabular}{l|cccccc}
\hline
Model &
SSIM $\uparrow$ &
IS $\uparrow$ &
DS $\uparrow$ &
PCKh $\uparrow$ &
\begin{tabular}[c]{@{}c@{}}LPIPS $\downarrow$\\ (VGG)\end{tabular} &
\begin{tabular}[c]{@{}c@{}}LPIPS $\downarrow$\\ (SqzNet)\end{tabular} \\ \hline
Baseline & 0.657 & \textbf{3.667} & 0.902 & 0.46 & 0.338 & 0.260 \\
HR only  & 0.825 & 3.271 & 0.954 & 0.96 & 0.154 & 0.088 \\
LR only  & 0.840 & 3.326 & 0.966 & 0.96 & 0.131 & 0.068 \\
Full     & \textbf{0.845} & 3.351 & \textbf{0.968} & \textbf{0.97} & \textbf{0.124} & \textbf{0.064} \\ \hline
Real & 1.000 & 3.687 & 0.970 & 1.00 & 0.000 & 0.000 \\ \hline
\end{tabular}%
}
\end{table}

We perform an extensive set of ablation experiments to optimize our generation pipeline. The ablation experiments and the observations are briefly discussed below.\\
\textbf{Feature representation during clustering:} As mentioned in Sec. \ref{sec:method_stage2}, we use 512-dimensional VGG-encoded features to guide the refinement process. To evaluate the effectiveness of VGG features in the proposed refinement strategy, we consider the raw pixel features in the ablation study by converting the input image into a feature vector. The conversion process downscales (nearest-neighbor interpolation) the original $176 \times 256$ images to $22 \times 32$, keeping the aspect ratio intact, followed by flattening to a 704-dimensional feature vector. We evaluate both the feature representation techniques for different numbers of clusters $(K=8, \; 16, \; 32, \; 64)$. As shown in Tables \ref{tab:ablation_clustering_vgg} \& \ref{tab:ablation_clustering_pixel}, for a particular value of cluster numbers $K$, VGG-encoded feature representation outperforms the raw pixel-based representation in average similarity score of top retrievals. As shown in Fig. \ref{fig:ablation_clustering}, our strategy uses similarity score-based ranking for both genders. The VGG feature-based clustering provides a better resemblance between the query and retrieved semantic maps. From our ablation study, we find $K=8$ works best for our data.\\
\textbf{Attention flow:} Attention in the rendering network plays an important role in the generated image quality. To check the attention mechanism's contribution and the optimal attention strategy, we explore four different settings. In the first setting (\textbf{Baseline}), we remove all attention operations and depth-wise concatenate $I^E_4$ with $H^E_4$. The concatenated feature space is passed through the decoder block. As shown in Table \ref{tab:ablation_rendering}, the Baseline model performs worst among all variants. We consider only one attention pathway in the rendering network in the second and third ablation settings. In the second variant (\textbf{HR only}), the attention operation is performed at the highest feature resolution only (just before the decoder block $D_4$). Similarly, in the third variant (\textbf{LR only}), the attention operation is performed at the lowest feature resolution only (just before the decoder block $D_1$). In the final settings (\textbf{Full}), we use the proposed attention mechanism as shown in Fig. \ref{fig:architecture} and described in Sec. \ref{sec:method_stage3}. We train and evaluate all four variants on the same dataset splits while keeping all experimental conditions the same, as noted in Sec. \ref{sec:experimental_setup}. We show the evaluated metrics in Table \ref{tab:ablation_rendering} along with qualitative results in Fig. \ref{fig:ablation_rendering}. We conclude from the analytical and visual results that the proposed attention mechanism provides the best generation performance.\\
\textbf{Refinement:} We show the efficacy of the data-driven refinement on the final generation in Fig. \ref{fig:ablation_refinement} by comparing the rendered scene with and without applying the refinement strategy.

\section{Limitations}\label{sec:limitations}
\begin{figure}[t]
  \centering
  \includegraphics[width=\linewidth]{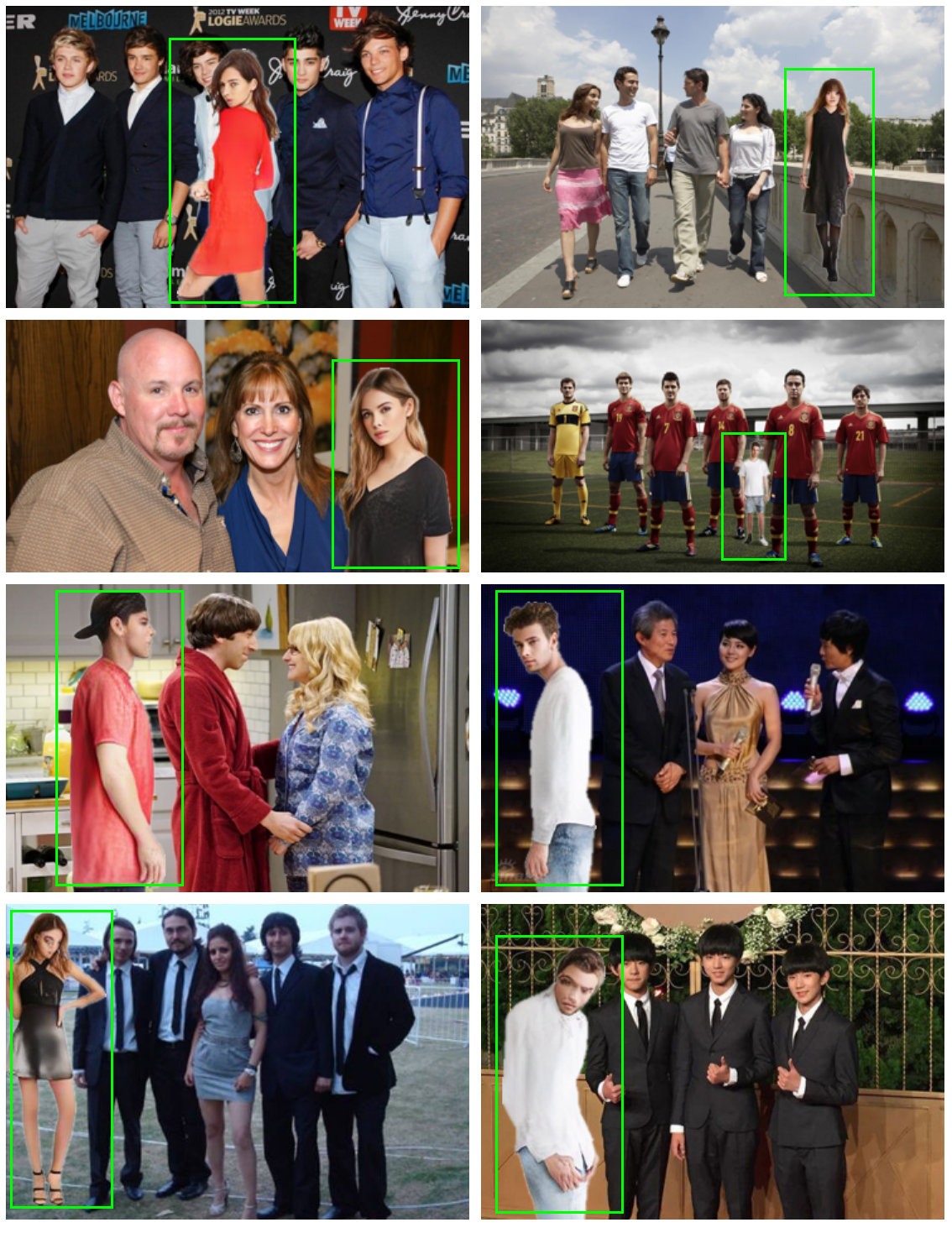}
  \caption{Limitations of the proposed method. (From top to bottom) \textbf{First row:} Positional inconsistency. \textbf{Second row:} Scale inconsistency. \textbf{Third row:} Contextual inconsistency. \textbf{Fourth row:} Rendering inconsistency.}
  \label{fig:limitations}
\end{figure}

Although the proposed method can produce high-quality, visually appealing results for a wide range of complex natural scenes, there are a few occasions when the technique fails to generate a realistic outcome. Due to a disentangled multi-stage approach, these limiting cases may occur from different pipeline components. In our method, coarse generation in stage 1 provides the spatial location and scale of the target person. Therefore, wrong inference in this step leads to a misinterpretation of the position and scale in the final target. The refined semantic target map is retrieved from the pre-partitioned clusters based on encoded features of the coarse semantic map in stage 2. Consequently, an extremely rough generation in stage 1 or a misclassified outlier during clustering in stage 2 can lead to a generated person that does not blend well with the existing persons in the scene. Finally, due to a supervised approach of training the renderer in stage 3, the appearance attribute transfer often struggles to generate high-quality outputs for both imbalanced and unconventional target poses. We show some of these limiting cases in Fig. \ref{fig:limitations}.

\section{Conclusions}\label{sec:conclusions}
In this work, we propose a novel technique for scene-aware person image synthesis by conditioning the generative process on the global context. The method is divided into three independent stages for a concise focus on individual subtasks. First, we use a coarse generation network based on the existing Pix2PixHD architecture to estimate the target person's spatial and pose attributes. While the spatial characteristics in the initial semantic map provide sufficient geometric information for the target, the semantic map itself does not preserve enough label group correctness, leading to improper attribute transfer in the rendering stage. We mitigate this issue through a data-driven distillation of the coarse semantic map by selecting candidate maps from a clustered knowledge base using a similarity score-based ranking. Finally, the appearance attributes from the exemplar are transferred to the selected candidate map using a generative renderer. The rendered instance is then injected into the original scene using the geometric information obtained during coarse generation. In our experiments, we achieve highly detailed realistic visual outcomes, which are further supported by relevant analytical evaluations. We also discuss an extensive ablation study and the limitations of our approach. We believe investigating a better way to model the global scene context and a robust end-to-end approach to the problem will benefit the potential future applications of the proposed method.

{\small
\bibliographystyle{ieee_fullname}
\bibliography{references}
}

\clearpage

\renewcommand{\thesection}{S\arabic{section}}
\renewcommand{\thefigure}{S\arabic{figure}}
\setcounter{section}{0}
\setcounter{figure}{0}

\onecolumn

\section*{\Large \textbf{Supplementary Material}}

\section{Additional results}

\begin{figure*}[h]
  \centering
  \includegraphics[width=\textwidth]{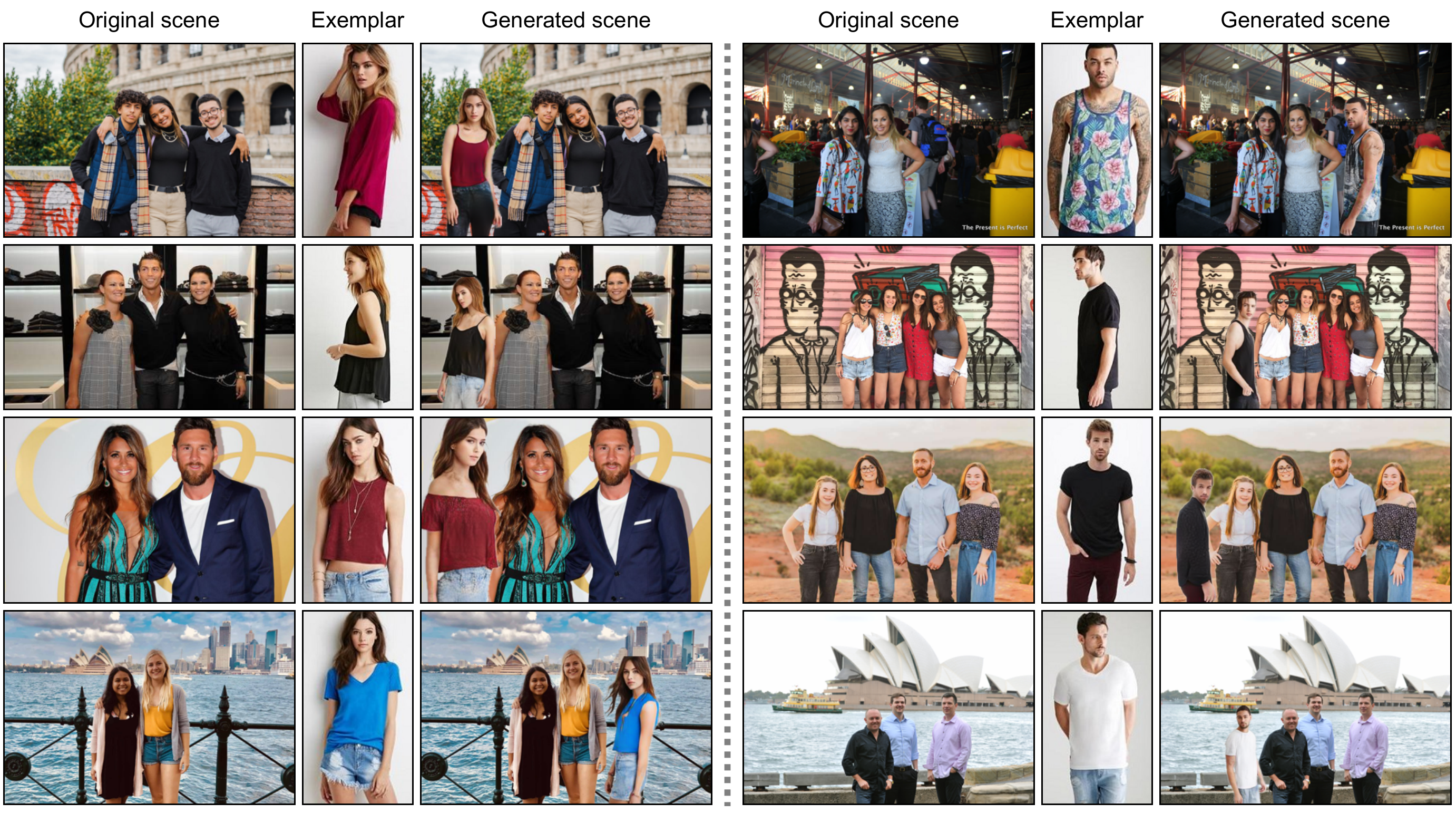}
  \caption{Qualitative results generated by the proposed method. Each set of examples shows -- the original scene (\textbf{left}), an exemplar of the target person (\textbf{middle}), and the final generated scene (\textbf{right}).}
  \label{fig:result_main}
\end{figure*}

In Fig. \ref{fig:result_main}, we show some additional qualitative results generated by the proposed method as supplemental to the main manuscript. However, in this case, the scene images are randomly collected from the internet as opposed to the main paper where we have shown the results on scenes obtained from the LV-MHP-v1 dataset \cite{li2017multiple}. We estimate the semantic maps corresponding to the existing persons in a scene image using a self-correcting human parser \cite{li2020self} pretrained on the ATR dataset \cite{liang2015deep}. The remaining steps to generate the final rendered scene follow an identical pipeline in sequence, as discussed in the main paper. From these results, we conclude that the proposed technique extends well beyond the experimental setup and can adapt to a wide range of complex natural scenes.

\section{Appearance diversity in refinement}

\begin{figure*}[t]
  \centering
  \includegraphics[width=\textwidth]{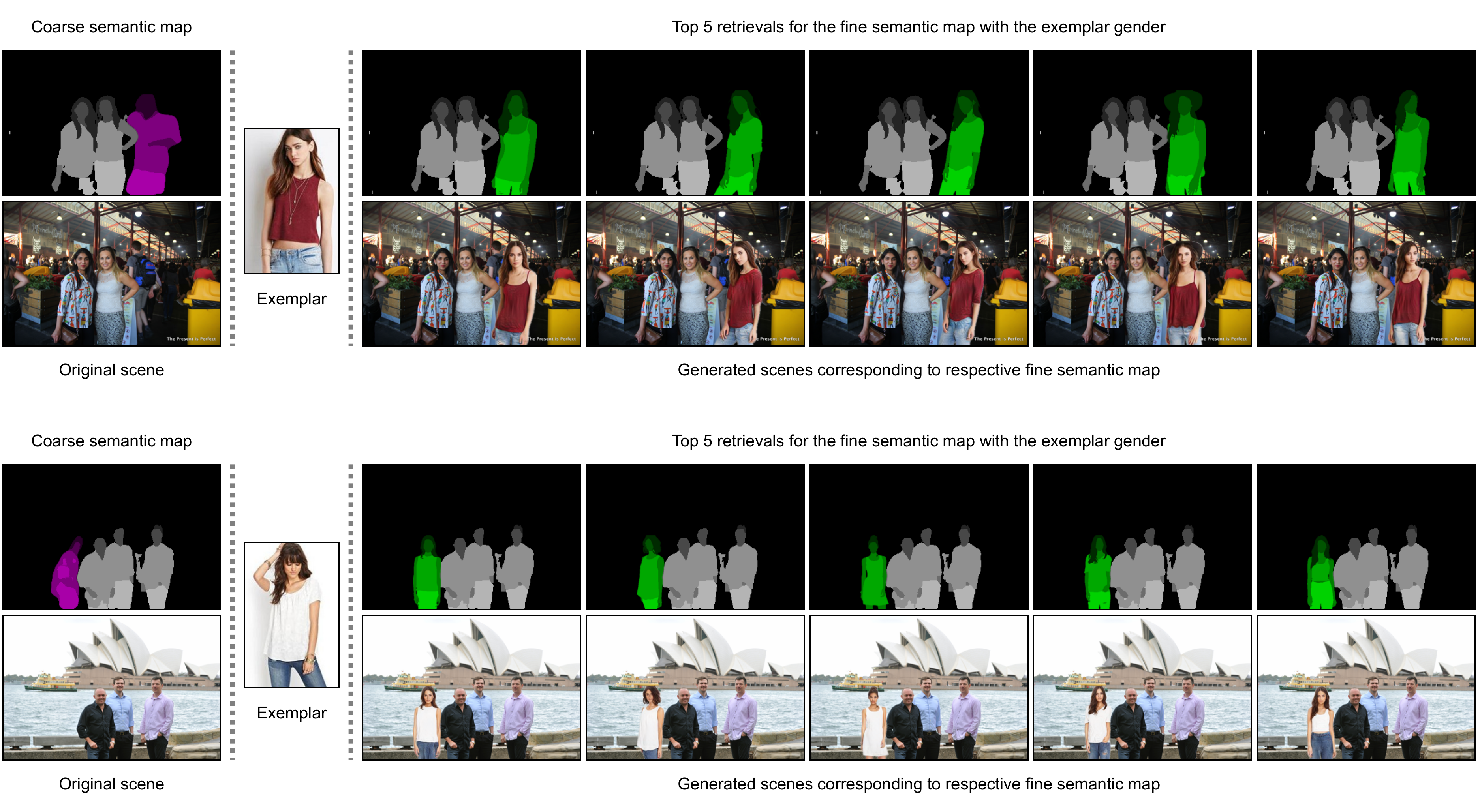}
  \caption{Qualitative results of unconstrained appearance diversity with data-driven refinement. \textbf{Top row:} Coarse estimation (\textcolor{magenta}{in purple}) followed by top 5 refined estimations (\textcolor{green}{in green}). \textbf{Bottom row:} Original scene followed by generated scenes from respective fine semantic maps. An exemplar of the target person provides gender information in the refinement query, and appearance attributes to the renderer.}
  \label{fig:result_top5}
\end{figure*}

We refine the initial coarse estimation of the target semantic map by retrieving closely aligned samples from pre-partitioned clusters using a cosine similarity-based ranking in the encoded latent space. Besides refining the coarse target semantic map, this data-driven strategy explicitly provides a way to achieve unconstrained appearance diversity for the synthesized person. In Fig. \ref{fig:result_top5}, we demonstrate a few examples to illustrate this idea by rendering the target person with the top 5 retrieved semantic maps from the refinement stage.

\section{Appearance control in rendering}

\begin{figure*}[t]
  \centering
  \includegraphics[width=\textwidth]{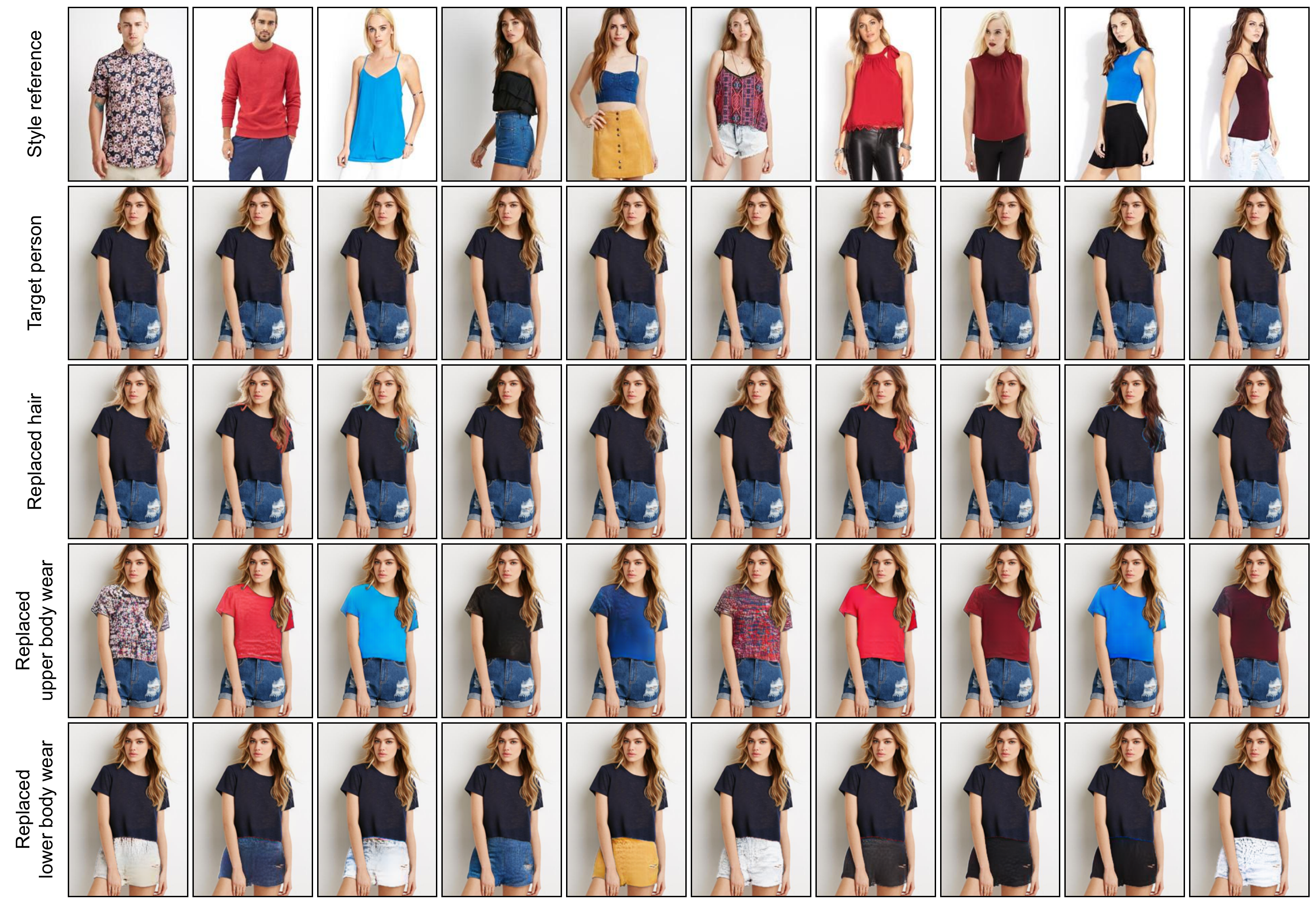}
  \caption{Qualitative results of appearance control in rendered person. (From top to bottom) \textbf{First row:} A person as the style reference. \textbf{Second row:} Target person. \textbf{Third row:} Target person with replaced hair as the reference. \textbf{Fourth row:} Target person with replaced upper body wear as the reference. \textbf{Fifth row:} Target person with replaced lower body wear as the reference.}
  \label{fig:result_swap1}
\end{figure*}

\begin{figure*}[t]
  \centering
  \includegraphics[width=\textwidth]{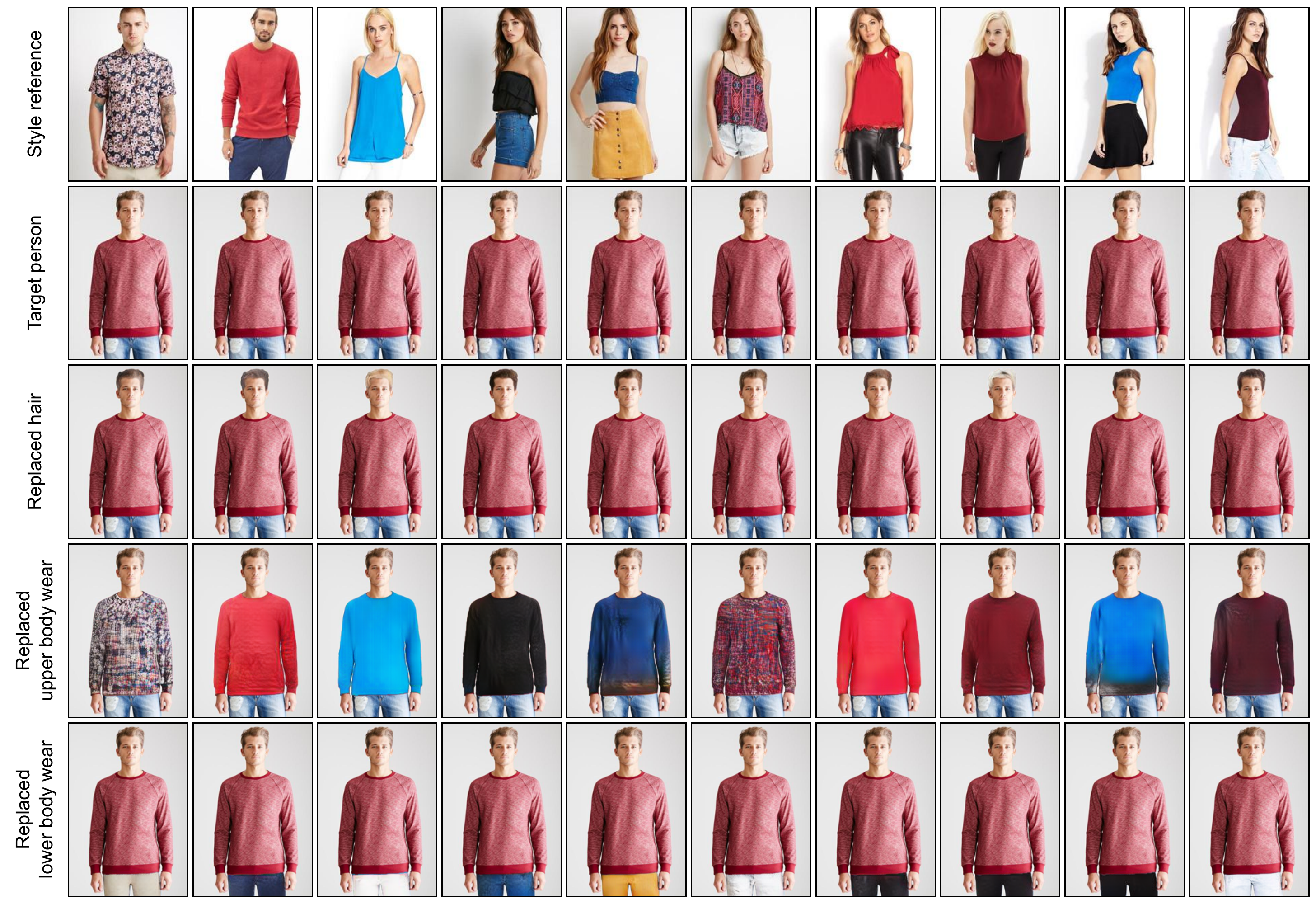}
  \caption{Qualitative results of appearance control in rendered person. (From top to bottom) \textbf{First row:} A person as the style reference. \textbf{Second row:} Target person. \textbf{Third row:} Target person with replaced hair as the reference. \textbf{Fourth row:} Target person with replaced upper body wear as the reference. \textbf{Fifth row:} Target person with replaced lower body wear as the reference.}
  \label{fig:result_swap2}
\end{figure*}

In the proposed method, we render the target person by transferring appearance attributes from an exemplar to the refined target semantic map. Different label groups in the semantic map can define precise segmentation masks for respective body parts. Therefore, it provides an implicit way to manipulate the appearance of various body regions of the target person by using bitwise operations. Mathematically, the modified target image $\hat{I}$ is given by
\begin{equation*}
  \hat{I} = \left[ M \; \odot \; \mathcal{G} \; (I_{style}, \; (H_{style}, \; H_{target})) \right] \; \oplus \; \left[ (1 - M) \; \odot \; I_{target} \right]
\end{equation*}
where $I_{style}$ and $I_{target}$ denote the style reference and the target person, respectively; $H_{style}$ and $H_{target}$ denote the heatmap representations for the semantic maps of the style reference and the target person, respectively; $M$ defines a binary segmentation mask for the body part being manipulated; $\mathcal{G}$ is the generator; $\odot$ and $\oplus$ are the bitwise multiplication and addition operations, respectively. In Fig. \ref{fig:result_swap1} \& \ref{fig:result_swap2}, we illustrate this idea of appearance control using samples from the DeepFashion dataset \cite{liu2016deepfashion}.

\end{document}